\documentclass{article} 
\usepackage{styleslatex,times}
\usepackage{fancyhdr}
\usepackage{natbib}
 \pdfoutput=1
\usepackage[utf8]{inputenc}
\usepackage[arabic, USenglish]{babel}

\usepackage{hyperref}
\usepackage{url}
\usepackage {graphicx} 
\graphicspath { {images/} }

\title{Towards A Sign Language Gloss Representation Of Modern Standard Arabic}

\author{Salma El Anigri \& M. Majid Himmi \\
LIMIARF Laboratory\\
Faculty of Sciences\\
Mohammed V University\\
Rabat, Morocco\\
\texttt{salmaelanigri@gmail.com} \\
\texttt{himmi.fsr@gmail.com} \\
\And
Abdelhak Mahmoudi \\
LIMIARF Laboratory\\
Ecole Normale Supèrieure\\
Mohammed V University\\
Rabat, Morocco\\
\texttt{abdelhak.mahmoudi@um5.ac.ma} \\
}

\iclrfinalcopy 
\begin{document}
\maketitle
 \begin{abstract}
Over 5\% of the world’s population (466 million people) has disabling hearing loss. 4 million are children. They can be hard of hearing or deaf. Deaf people mostly have profound hearing loss, which implies very little or no hearing. Over the world, deaf people often communicate using a sign language with gestures of both hands and facial expressions. The Sign Language is a full-fledged natural language with its own grammar and lexicon. Therefore, there is a need for translation models from and to sign languages. In this work, we are interested in the translation of Modern Standard Arabic (MSAr) into Sign Language. We generated a gloss representation from MSAr that extracts the features mandatory for the generation of animation signs. Our approach locates the most pertinent features that maintain the meaning of the input Arabic sentence.
\end{abstract}
\section{Introduction}
The main impacts of deaf is on the individual’s ability to communicate with others in addition to the emotional feelings of loneliness and isolation in society. Consequently, they can not equally access public services, mostly education and health and have not equal rights in participating at the active and democratic life. This leads to a negative impact in their lives and the lives of the people surrounding them.

Over the world, deaf people often communicate using a sign language with gestures of both hands and facial expressions. The Sign Language is a full-fledged natural language with its own grammar and lexicon. However, sign languages are not universal where each natural language have its corresponding sign language. As research in machine translation from one language to another increased in recent years; mainly due to advances in deep learning field; research in translation from and to sign languages increased as well. 

\begin{figure}[htp]
\centering
\includegraphics[width=4cm]{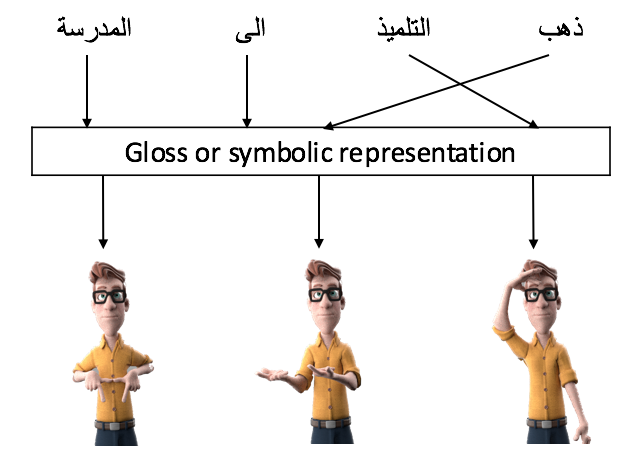}
\caption {Translation of a natural language to a sign animation.}
\label{fig:Figure1}
\end{figure}

Typically, a machine translation system from an input language to a sign language can be divided into two stages. The first stage consists of finding a mapping between the text of the input language and a sign representation. In the second stage, the sign representation is matched to a visual sign using an animated avatar (Fig. \ref{fig:Figure1}). The representation of the sign could be textual; one talk about the gloss representation; or symbolic such as Hamburg representation \cite{2004_hamburg}.

In the literature, in the best of our knowledge, there is very few research on the translation of Modern Standard Arabic (MSAr) into Sign Language. This is mainly due to the morphological and grammatical complexity of the MSAr and to the lack of data. In this work, we aim to generate a gloss representation from MSAr that extracts the features mandatory for the generation of the animation signs.

\section{Problem Statement}
Unlike the majority of the natural languages, the Modern Standard Arabic Language is written from right to left. Furthermore, it contains ambiguities in many aspects, such as problems to identify the right function or meaning of a word, the absence of capital letters which complicates the differentiation between proper nouns, adjectives, nouns, abbreviations and acronyms \cite{farghaly_2009}. In statistical Arabic translation systems, dealing with those ambiguities were shown to have significant effects \cite{habash_2019}. This observation remains true when one want to translate from MSAr to a sign language gloss representation \cite{manny_2016, aouiti_translation_2018, luqman_automatic_2019}. In this work, we want to generate gloss representation from MSAr while taking into consideration both the pre-processing phase and the usability of such a representation in the animation stage. 

\section{Approach and Results}

This study aims to extract a set of relevant features while preserving the meaning of the input phrase, to perform understandable sign animations by the deaf. 

This can be achieved by extracting phrase tense and forms using the lemmas and the part of speech tags of each token. Furthermore, words expressing emphasis like "very (\textRL{جدا})" and "often (\textRL{مرارا})" have to be recognised. On one side, depending on the part of speech tags, one can capture the past, present or future tense of the phrase. By default, a verb is always in its past tense. In the case where it begins with (\textRL{س}), it switches to the future tense. Otherwise, if it begins with \textRL{ت، ن، ي، أ}, the verb can be considered to be in the present tense. On the other side, a phrase form could be negative, interrogative or conditional. In the case of negative form, the tense can be captured using particles. Indeed, the negative particle "la (\textRL{لا})" and "layssa (\textRL{ليس})" indicates the present tense, "lan (\textRL{لن})", indicates the future tense and "lam (\textRL{لم}" indicates the past tense. In addition, usually, interrogative forms can be detected using the question mark (?) punctuation. Each interrogative form could be labeled depending the associated adverb ("is (\textRL{هل})", "who (\textRL{من})", "where (\textRL{أين})", "when (\textRL{متى}), "what (\textRL{ماذا})" and "how (\textRL{كيف})). Lastly, phrases in the conditional form contain the "if" token ("if (\textRL{إذا}) and "if (\textRL{لو})".

\begin{figure}[htp]
\centering
\includegraphics[width=10cm]{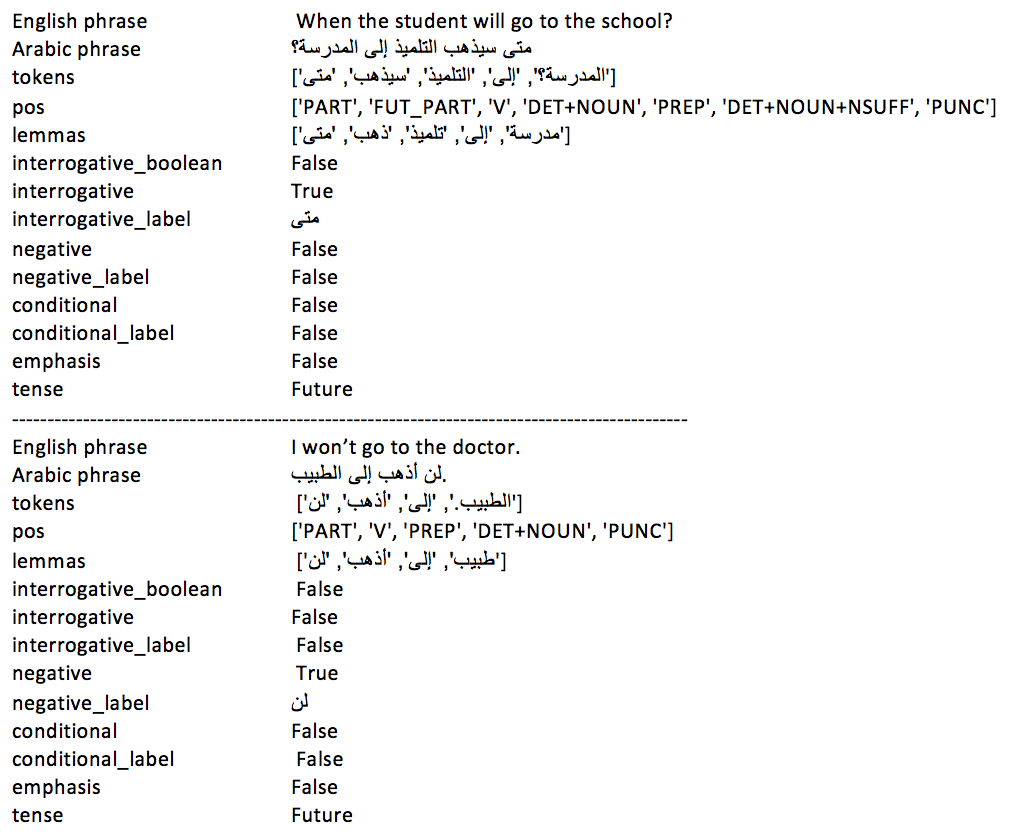}
\caption {Two examples of the gloss representation of two Arabic sentences. Top, interrogative form and down, negative form.}
\label{fig:Figure2}
\end{figure}

 For the extraction of features we were based on the work of \cite{aouiti_translation_2018}, especially the way they introduced the glossing rules by analogy to the ASL gloss in \cite{aouiti_arab_2015}. Our contribution can be seen in the way we extracted each feature; we combined more grammar rules of the Arabic language. Such as, To detect the future tense, we see if the  part of speech tags contains a future particle  sin " (\textRL{س })", or the particle lan " (\textRL{لن })" is identified.
 Figure \ref{fig:Figure2} shows two gloss representations based on those established rules.

\section{Conclusion and Perspectives}
In this work, we were interested in the first stage of the translation from Modern Standard Arabic to sign language animation that is generating a sign gloss representation. We identified a set of rules mandatory for the sign language animation stage and performed the generation taking into account the pre-processing proven to have significant effects on the translation systems. The presented results are promising but far from well satisfying all the mandatory rules. For example, the gloss representation we proposed don't take into account the pronouns to express genders, plurals or singulars. Furthermore, the named entities expressing locations, people or times are not included. In upcoming work, we aim to tackle these problems.

\subsubsection*{Acknowledgments}
This work has been funded by Artificial Intelligence 4 Development (AI4D) programme as part of the 1st Call for Project Proposals on Artificial Intelligence 4 Development Technologies.

\bibliography{final_version}

\begin{thebibliography}{7}
\providecommand{\natexlab}[1]{#1}
\providecommand{\url}[1]{\texttt{#1}}
\expandafter\ifx\csname urlstyle\endcsname\relax
  \providecommand{\doi}[1]{doi: #1}\else
  \providecommand{\doi}{doi: \begingroup \urlstyle{rm}\Url}\fi

\bibitem[Aouiti \& Jemni(2018)Aouiti and Jemni]{aouiti_translation_2018}
Nadia Aouiti and Mohamed Jemni.
\newblock Translation {System} from {Arabic} {Text} to {Arabic} {Sign}
  {Language}.
\newblock \emph{Journal of Applied Intelligent System}, 3:\penalty0 57--70,
  2018.

\bibitem[Aouiti et~al.(2015)Aouiti, Jemni, and Semreen]{aouiti_arab_2015}
Nadia Aouiti, Mohamed Jemni, and Sameer Semreen.
\newblock Arab gloss annotation system for {Arabic} {Sign} {Language}.
\newblock In \emph{2015 5th {International} {Conference} on {Information} \&
  {Communication} {Technology} and {Accessibility} ({ICTA})}, pp.\  1--6. IEEE,
  2015.

\bibitem[Farghaly \& Shaalan(2009)Farghaly and Shaalan]{farghaly_2009}
Ali Farghaly and Khaled Shaalan.
\newblock Arabic natural language processing: Challenges and solutions.
\newblock \emph{ACM Transactions on Asian Language Information Processing
  (TALIP)}, 8\penalty0 (4), December 2009.
\newblock ISSN 1530-0226.
\newblock \doi{10.1145/1644879.1644881}.
\newblock URL \url{https://doi.org/10.1145/1644879.1644881}.

\bibitem[Hanke(2004)]{2004_hamburg}
T.~Hanke.
\newblock Hamnosys - representing sign language data in language resources and
  language processing contexts.
\newblock In \emph{Workshop proceedings : Representation and processing of sign
  languages}, pp.\  1--6. IEEE, 2004.

\bibitem[Luqman \& Mahmoud(2019)Luqman and Mahmoud]{luqman_automatic_2019}
Hamzah Luqman and Sabri~A. Mahmoud.
\newblock Automatic translation of {Arabic} text-to-{Arabic} sign language.
\newblock \emph{Univ Access Inf Soc}, 18:\penalty0 939--951, 2019.

\bibitem[Oudah et~al.(2019)Oudah, Almahairi, and Habash]{habash_2019}
Mai Oudah, Amjad Almahairi, and Nizar Habash.
\newblock The impact of preprocessing on {A}rabic-{E}nglish statistical and
  neural machine translation.
\newblock In \emph{Proceedings of Machine Translation Summit XVII Volume 1:
  Research Track}, pp.\  214--221, Dublin, Ireland, August 2019. European
  Association for Machine Translation.
\newblock URL \url{https://www.aclweb.org/anthology/W19-6621}.

\bibitem[Rayner et~al.(2016)Rayner, Armando, Bouillon, Ebling, Gerlach, Halimi,
  Strasly, and Tsourakis]{manny_2016}
Manny Rayner, Alejandro Armando, Pierrette Bouillon, Sarah Ebling, Johanna
  Gerlach, Sonia Halimi, Irene Strasly, and Nikos Tsourakis.
\newblock Helping domain experts build phrasal speech translation systems.
\newblock In Jos{\'e}~F. Quesada, Francisco-Jes{\'u}s Mart{\'i}n~Mateos, and
  Teresa Lopez-Soto (eds.), \emph{Future and Emergent Trends in Language
  Technology}, pp.\  41--52, Cham, 2016. Springer International Publishing.
\newblock ISBN 978-3-319-33500-1.

\end{thebibliography}

\bibliographystyle{refsty}


\end{document}